\documentclass{article}

\usepackage[final,nonatbib]{nips_2017}

\usepackage[utf8]{inputenc} % allow utf-8 input
\usepackage[T1]{fontenc}    % use 8-bit T1 fonts
\usepackage{booktabs,ragged2e,amsfonts,tabularx,hyperref,url,times,
courier,csquotes,dblfloatfix,setspace,ifthen,amsmath, amsthm, amssymb, wrapfig,graphicx,amsfonts,multirow,subfig,todonotes,algorithmic,times,algorithm}

\newcommand*\samethanks[1][\value{footnote}]{\footnotemark[#1]}
\author{
  Ershad Banijamali$^{1}$\thanks{equal contribution} ,
  Amir-Hossein Karimi$^{1}$\samethanks[1] ,
  Alexander Wong$^2$,
  Ali Ghodsi$^3$ 
   \\  
  $^1$School of Computer Science,  University of Waterloo \\
  $^2$Systems Design Engineering, University of Waterloo \\
  $^3$Department of Statistics and Actuarial Science, University of Waterloo \\
  \texttt{\{a6karimi, sbanijam, a28wong, aghodsib\}@uwaterloo.ca} \\
}

\title{JADE: Joint Autoencoders for Dis-Entanglement}

\usepackage{color,soul}
\usepackage{hhline}
\newcommand{\oo}[2]{#1{\tiny \pm #2}}
\newcolumntype{L}[1]{>{\raggedright\let\newline\\\arraybackslash\hspace{0pt}}m{#1}}
\newcolumntype{C}[1]{>{\centering\let\newline\\\arraybackslash\hspace{0pt}}m{#1}}
\newcolumntype{R}[1]{>{\raggedleft\let\newline\\\arraybackslash\hspace{0pt}}m{#1}}

\begin{document}

\maketitle

\begin{abstract}
The problem of feature disentanglement has been explored in the literature, for the purpose of image and video processing and text analysis. State-of-the-art methods for disentangling feature representations rely on the presence of many labeled samples. In this work, we present a novel method for disentangling factors of variation in data-scarce regimes. Specifically, we explore the application of feature disentangling for the problem of supervised classification in a setting where few labeled samples exist, and there are no unlabeled samples for use in unsupervised training. Instead, a similar datasets exists which shares at least one direction of variation with the sample-constrained datasets. We train our model end-to-end using the framework of variational autoencoders and are able to experimentally demonstrate that using an auxiliary dataset with similar variation factors contribute positively to classification performance, yielding competitive results with the state-of-the-art in unsupervised learning.

\end{abstract}

%% %% %% %% %% %% %% %% %% %% %% %% %% %% %% %% %% %% %% %% %% %% %% %% %% %% %% %% %% %% 
%% %%          INTRODUCTION                                                         %% %%
%% %% %% %% %% %% %% %% %% %% %% %% %% %% %% %% %% %% %% %% %% %% %% %% %% %% %% %% %% %% 
\vspace{-.2cm}
\section{Introduction}
\vspace{-.2cm}
In machine learning, samples in a dataset originate via complicated processes driven by a number of underlying factors. Individual factors lead to independent directions of variations in the observed samples,  while the accumulation of factors give rise to the rich structure characteristic of these datasets. The underlying factors often interact in complicated and unpredictable ways, and appear tightly \textit{entangled} in the raw data. Being able to tease apart the effect of underlying factors is a fundamental challenge in understanding these datasets.

For instance, a dataset containing images of natural scenery may be subject to variation in lighting conditions, camera elevation, and the appearance of the scene itself. Controlling and restraining variation at data acquisition time is difficult, and limits the number of acceptable samples in the dataset. On the other hand, capturing annotations for every direction of variation is time-consuming and infeasible. Therefore, designing methods that automatically learn to separate out underlying factors (known and unknown) is relevant for many applications in machine learning.

One area that has enjoyed tremendous success for separating factors of variation is supervised learning. The representations learned here aim to satisfy a specific task that is driven by the explicit labels in the dataset. Therefore, these representations are invariant to factors of variation that are uninformative for solving the task at hand. For example, when identifying individuals in a school yearbook, the identity of the person is paramount compared to their facial expression. Hence, a simple method that simply discards the irrelevant variation in expression will perform really well. Learning invariant representations, however, require many samples and comes at the cost of needing to train a new model for a closely related task that depends on an alternative direction of variation. It would seem reasonable then to desire a strategy that captures all directions of variation in a single model in a \textit{disentangled} manner allowing one to infer all factors for a given sample in the absence of labels for each factor.

% \hl{Furthermore, models that preserve all directions of variation often times perform better than their feature-dropping (?) counterparts, as they are able to better generalize (??) and generate new realistic samples to bootstrap (??) / improve the model.} 

Current state-of-the-art strategies for disentangling factors of variation mostly fall victim to the challenges in deep learning and rely on the presence of abundant data samples. In \cite{kulkarni2015deep}, the authors were able to accurately separate out lighting, pose, and shape while sampling seemingly unlimitedly from an auxiliary generative model that creates samples with different variations. The results presented in \cite{reed2014learning, mathieu2016disentangling} also build upon datasets containing often hundreds of thousands of samples. Whereas \cite{kingma2014semi,siddharth2017learning} use very few samples in their training process, these methods are semi-supervised and have access to unlabeled samples from the same dataset following the same statistical distribution.

In this work, we explore classification in a data-scarce scenario where not only are there few labeled samples available, there are also no unlabeled samples from which one could perform semi-supervised training.  These situations commonly arise in medical imaging datasets, e.g., pancreatic cancer MRI images are scarce whereas breast cancer MRI images are abundant (\cite{greenspan2016guest} and references therein). In such a situation, we ask whether one can employ a secondary dataset, with many samples, similar content, but different style, to improve the performance of a benchmark classification model. What remains to be demonstrated is how to learn good intermediate representations that can be shared across tasks and use the disentanglement process of the secondary dataset to effectively disentangle the factors of variation in the primary dataset of interest. Essentially we are entangling together the feature disentangling of two similar datasets. This is the focus of the work below.

%% %% %% %% %% %% %% %% %% %% %% %% %% %% %% %% %% %% %% %% %% %% %% %% %% %% %% %% %% %% 
%% %%          MODEL DESCRIPTION                                                    %% %%
%% %% %% %% %% %% %% %% %% %% %% %% %% %% %% %% %% %% %% %% %% %% %% %% %% %% %% %% %% %% 
\vspace{-.2cm}
\section{Model Description}
\vspace{-.2cm}
In this work, we consider a situation where we are given a labeled dataset, $X$, with limited number of points. We denote the label variable by $\ell$. We also have access to another dataset $Y$ with a larger number of points that share the same categories as $Y$. However, the underlying distribution of the datasets are different. Let us denote the distribution for $X$ and $Y$ by $p(x)$ and $p(y)$, respectively. Suppose that our goal is to classify unseen data points that come from $p(x)$, i.e. to maximize $p(\ell|x)$. Building a classifier that simply uses $X$ can lead to low accuracy and overfitting, due to its small size. Therefore we want to leverage the information of $Y$ about the label variable and build a model that can classify the points from $p(x)$ with higher accuracy.

\begin{figure}[!b]
\vspace{-1cm}
    \centering
    \subfloat[]{{\includegraphics[trim = 0mm 0mm -10mm 0mm,height=4cm]{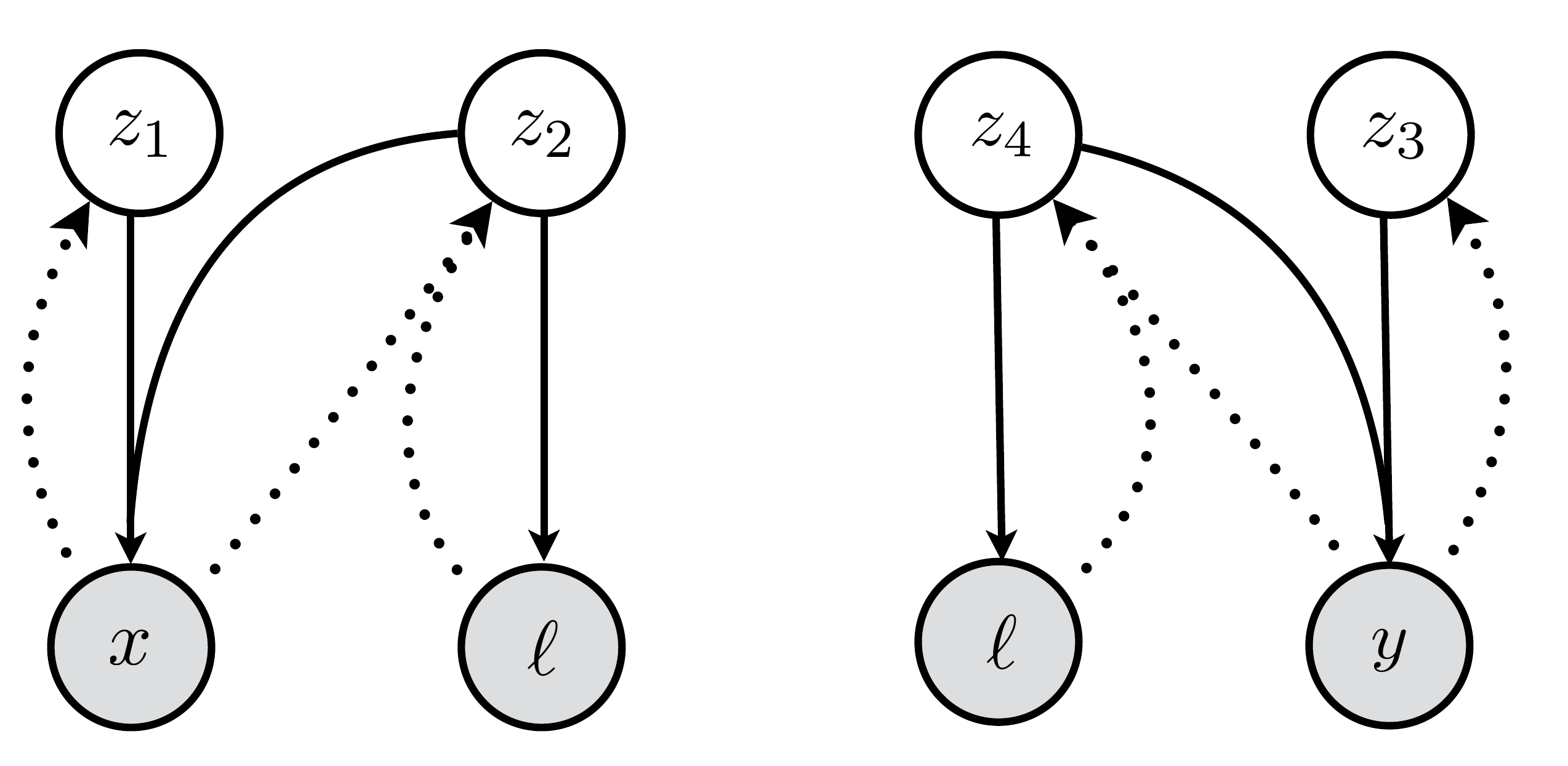}  
    \label{fig:gm}}}
	\subfloat[]{{\includegraphics[trim = -10mm 0mm 0mm 0mm,height=4cm]{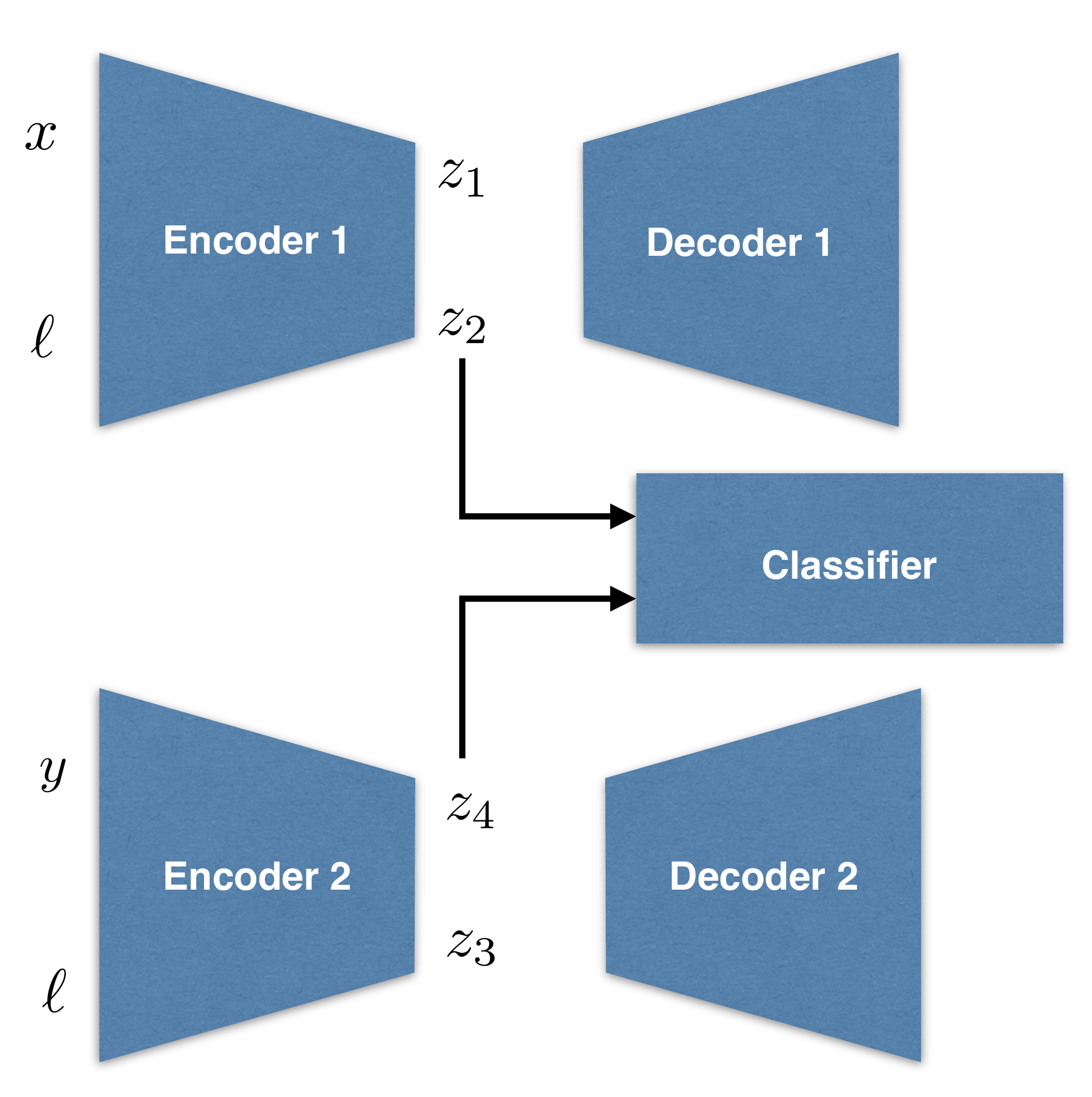}
    \label{fig:networks}}} 
    \vspace{-.2cm}
    \caption{(a) Graphical models of the method. (b) Network structure of the method}
\end{figure}

Our approach to address this problem is to disentangle the features in $X$ and $Y$ that contribute in predicting the label variable (i.e., content) from the features that contribute to the style of $X$ and $Y$. Consider the graphical model in Fig. \ref{fig:gm}. We assume there are two pairs of latent variables that describe each of $x$ and $y$. Based on this figure, suppose that $z_1$ and $z_2$ generate samples in dataset $X$ and $z_3$ and $z_4$ generated samples in dataset $Y$. If we assume that $z_2$ and $z_4$ are the latent variables that carry all the information about the label variable $\ell$ then $p(\ell|z_2) = p(\ell|z_4)$. Considering the same prior distributions over $z_2$ and $z_4$, i.e. $\mathcal{N}(0,I)$, we can guarantee the disentanglement of latent features by asserting that $p(z_2|\ell) = p(z_4|\ell)$. However, these posteriors are intractable. To approximate them we use the framework of variational inference where $p(z_2|\ell)$ and $p(z_4|\ell)$ are approximated by $q(z_2|x,\ell)$ and $q(z_4|y,\ell)$, respectively. Therefore, by matching these approximating distribution, we guarantee that only $z_2$ and $z_4$ carry information regarding the label $\ell$ (i.e., content) and therefore are disentangled from $z_1$ and $z_3$ respectively which represent style. 

All the conditional distributions on the graphical models in Fig. \ref{fig:gm} are parameterized by the neural networks depicted in Fig. \ref{fig:networks}. The joint model here builds on earlier work in \cite{rifai2012disentangling} where an autoencoder and a discriminator were trained in the framework of contractive discriminative analysis for semi-supervised learning. Here, we use the variational autoencoding \cite{kingma2013auto} approach to jointly train two networks that simultaneously extract shared discriminative features present in the primary and secondary datasets. This architecture is reminiscent of Domain Separation Networks \cite{bousmalis2016domain}. The proposed JADE model, however, focuses on a shared classifier for improved classification and joint disentanglement instead of a shared encoder and decoder.

The variational lower bound on the joint distribution of the observations is:
\begin{equation}
\begin{array}{l}
\log p(x,\ell) \geq \mathcal{L}(x,\ell) = \mathbb{E}_{\substack{q(z_1|x)\\q(z_2|x,\ell)}} \big[\log p(x|z_1,z_2) \big] + \mathbb{E}_{q(z_2|x,\ell)} \big[ \log p(\ell|z_2) \big] \\ 
\hspace{3.15cm} - \text{KL} \big(q(z_1|x) \parallel p(z_1) \big) - \text{KL} \big(q(z_2|x,\ell) \parallel p(z_2)) \big) \\ \\
\log p(y,\ell) \geq \mathcal{L}(y,\ell) = \mathbb{E}_{\substack{q(z_3|y)\\q(z_4|y\ell)}} \big[\log p(x|z_3,z_4) \big] + \mathbb{E}_{q(z_4|y,\ell)} \big[ \log p(\ell|z_4) \big] \\
\hspace{3.15cm} - \text{KL} \big(q(z_3|x) \parallel p(z_3) \big) - \text{KL} \big(q(z_4|x,\ell )\parallel p(z_4)) \big) \\
\end{array}
\end{equation}

We would like to maximize the sum over the above lower bounds. The approximating distributions are from exponential family (Gaussian) and to match them we assume that for the samples that are from the same class in the two datasets, we want to minimize  $\text{KL} \big ( q(z_2|x,\ell) \parallel q(z_4|y,\ell)\big)$. Given this condition, the overall objective of the model is: 
\begin{equation}
\max_{\Theta} \mathcal{L}(x,\ell) + \mathcal{L}(y,\ell) - \text{KL} \big ( q(z_2|x,\ell) \parallel q(z_4|y,\ell)\big)
\end{equation}
where $\Theta$ represents the entire parameter set of neural networks.

%% %% %% %% %% %% %% %% %% %% %% %% %% %% %% %% %% %% %% %% %% %% %% %% %% %% %% %% %% %% 
%% %%          EXPERIMENTS                                                          %% %%
%% %% %% %% %% %% %% %% %% %% %% %% %% %% %% %% %% %% %% %% %% %% %% %% %% %% %% %% %% %% 
\vspace{-.2cm}
\section{Experiments}
\vspace{-.2cm}
% \hl{We evaluate our framework along a number of different axes pertaining to its ability to learn disentangled
% representations} and perform supervised classification. 

\textbf{Datasets:} Our framework addresses the problem of performing supervised classification in data-scarce regimes where there exists a secondary dataset that has at least one direction of variation in common with the primary sample-constrained dataset. In our experiments we emulate this scenario with commonly used datasets such as MNIST \cite{lecun2010mnist} and SVHN \cite{netzer2011reading}. Because MNIST is relatively easier to learn, even with very few samples, we select SVHN as the sample-constrained primary dataset that is difficult to learn, and use the entirety of MNIST as the secondary dataset. These datasets differ in appearance and style: whereas MNIST is gray-scale and comes in $28 \times 28$ pixel images, SVHN has three color channels and comes in $32 \times 32$ pixel images. However, both datasets represent the same content (i.e., digit values) across different styles. This similarity in content of both datasets is what makes MNIST a good secondary dataset to boost SVHN's supervised classification performance.

\textbf{Model Comparison:} To evaluate the performance of our framework, we first develop a benchmark for supervised classification of SVHN. Here, we choose a relatively powerful convolutional neural network (CNN) architecture combined with a multi-layer perceptron (MLP) as the supervised classification model. The CNN architecture comprises of $4$ layers of $3 \times 3$ spatial convolutions ($\{64,96,64,8\}$ filters respectively) followed by ReLU and interspersed with $3$ layers of $2\times$ max-pooling. The MLP contains $3$ blocks of $500$-dimensional fully connected layers, followed by ReLU and Dropout ($p = 0.5$) layers \cite{srivastava2014dropout}. A $10$-dimensional bottleneck layer was placed in between the CNN and the MLP to encourage only important features from being retained. A final softmax layer is present at the end of the network for $10$-way classification. The loss for this model is measured using categorical cross-entropy. This architecture is referred to as \textit{single classifier} (i.e., benchmark). 

A simple extension of above setup is a model that jointly trains SVHN and MNIST on a shared MLP classifiers using features extracted from separate CNN feature extractors, one per dataset. The CNN used for SVHN and the MLP follow the same architecture as the benchmark above. The CNN architecture for MNIST comprises of $3$ layers of $3 \times 3$ spatial convolutions ($\{32,32,16\}$ filters respectively) followed by ReLU and interspersed with $3$ layers of $2\times$ max-pooling. A $10$-dimensional bottleneck layer was placed in between the CNN for MNIST and the shared MLP to capture the latent features of MNIST. Feature-extracted samples from both datasets are fed into the shared MLP in alternation and trained jointly. The loss of the system is the sum of the categorical cross-entropy losses for both datasets on the shared classifier. This setup is called \textit{paired classifier}. 

Finally, the proposed model (outlined in Fig. \ref{fig:networks}) extends upon the previous two methods by adding a decoder network to reconstruct the $10$-dimensional latent representations from each of the CNN feature extractors. To encourage disentanglement of features in the latent space, and to perform factor separation in a way that the MLP classifier is only given content-related features (i.e., digit values), we increase the size of the latent spaces from $10$ to $20$ dimensions. However, only $10$ of the latent dimensions resulting from each CNN are passed into the shared MLP, essentially keeping consistent with the previous method in terms of classifier capacity. All $20$ latent dimensions are used to reconstruct the inputs via a decoder that identically mirrors the corresponding CNN ($2\times$ up-sampling layers used in place of $2\times$ max-pooling). Losses are defined in Section 2. Due to the autoencoding structure of this model, we refer to it as \textit{JADE: Joint Autoencoders for Dis-Entanglement}.

% This same architecture is used when comparing with the performance of the proposed framework. Finally, the results are compared with the state-of-the-art in semi-supervised training of SVHN on limited samples \cite{kingma2014semi, siddharth2017learning}

% \textbf{Training Procedure:}
% All models were trained with the AdaM optimizer \cite{kingma2014adam} momentum-correction set to the default value. Models were trained for $300$ epochs on mini-batches of size $100$. Initial learning rates were set to $0.001$, and a scheduler divided the learning rate by $\sqrt{10}$ every $100$ epochs.

\begin{table}[!t]
  \centering
  \caption{\small Classification error rates for SVHN on limited data: $100$ samples per each class. Error rates calculated using the entirety of SVHN's test set. Results of our experiments are averaged over 3 runs. We observe improved SVHN classification performance without sacrificing near state-of-the-art performance on MNIST.}
  \begin{tabular}{l|| C{3.5cm} || C{3.5cm} }
	Method											& SVHN (1000 samples)	& MNIST (45K samples)	  		\\ \hhline{===}
    VAE (M1+M2) \cite{kingma2014semi}				& $\oo{36.02}{0.10}$ 	& -								\\ 
    Siddharth et al. \cite{siddharth2017learning}	& $\oo{28.71}{2.38}$ 	& -								\\ \hline
    Single Classifier (benchmark)					& $\oo{32.31}{1.56}$	& -								\\ 
    Paired Classifier								& $\oo{30.17}{2.77}$	& $\oo{0.82}{0.05}$ 			\\ 
    JADE (proposed)									& $\oo{29.08}{0.92}$	& $\oo{0.72}{0.03}$				\\ 
  \end{tabular}
  \label{table:results_summary}
  \vspace{-.5cm}
\end{table}

%% %% %% %% %% %% %% %% %% %% %% %% %% %% %% %% %% %% %% %% %% %% %% %% %% %% %% %% %% %% 
%% %%          DISCUSSION                                                           %% %%
%% %% %% %% %% %% %% %% %% %% %% %% %% %% %% %% %% %% %% %% %% %% %% %% %% %% %% %% %% %% 
\textbf{Discussion:} The results of our experiments have been presented in Table \ref{table:results_summary}. Here we compare the results of the single classifier (i.e., benchmark model), paired classifier, and proposed model (JADE) alongside those from Kingma et al. \cite{kingma2014semi} and Siddharth et al. \cite{siddharth2017learning}. It is worth pointing out that the former 3 models are trained only on $1000$ labeled sample from SVHN, whereas the cited models use the remainder of the SVHN training dataset in an unsupervised fashion. We, on the other hand, use all of the MNIST dataset to train the paired classifier in JADE.
\begin{figure}[!b]
\vspace{-.7cm}
\centering
\includegraphics[trim = 20mm 0mm 0mm 0mm,width = 10cm]{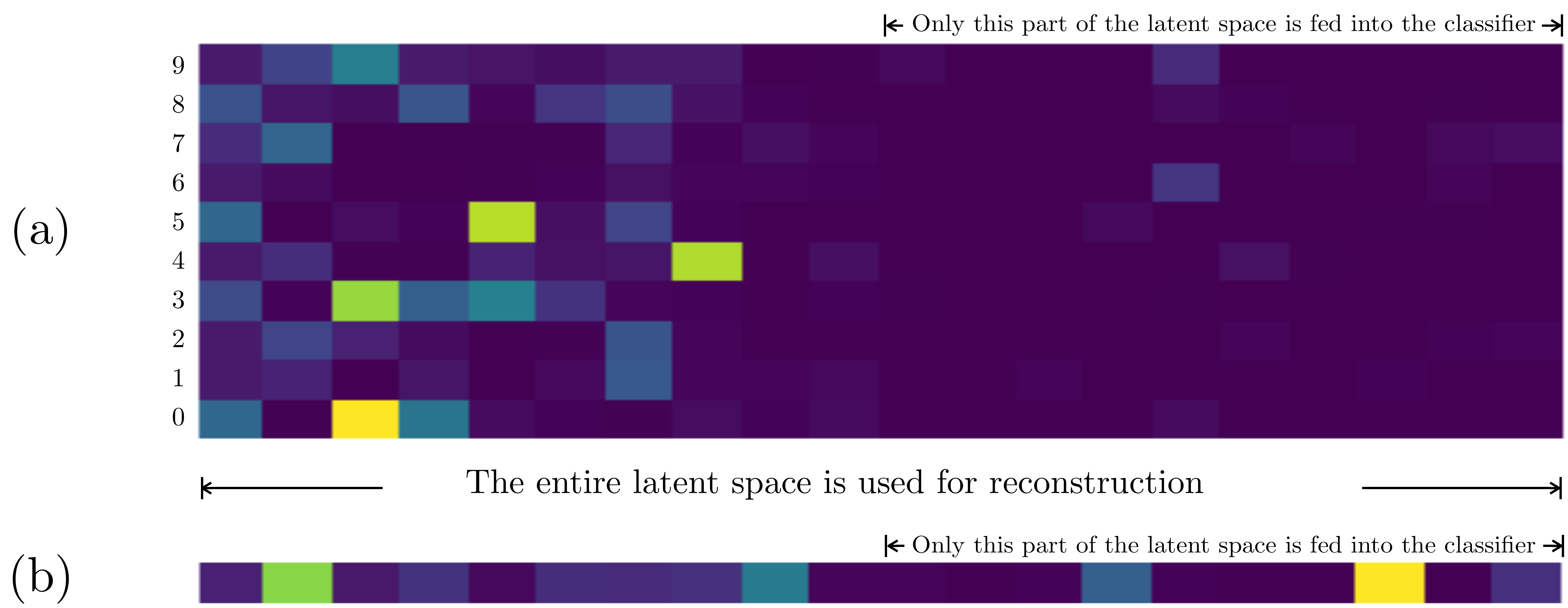}
\caption{
\small \textbf{(a) }variance normalized activations of latent space parameters, averaged over $500$ random samples from each of $10$ classes in SVHN; when content is fixed, the part of the latent space that feeds into the classifier exhibits weaker variance in activations compared to the part of the latent space that seemingly represents style over the $500$ samples. \textbf{(b)} variance normalized activations of latent space parameters for $2500$ random samples from SVHN spanning various style and content; all $20$ latent space parameters fire for random splits of the data.}
\vspace{-.6cm}
\end{figure}

These results demonstrate that when dealing with sample-constrained regimes without unlabeled samples, one can use a similar dataset with at least one shared direction of variation to improve classification performance. This can be seen when comparing the performance of a single classifier ($\oo{32.31}{01.56}$) with that of a paired classifier ($\oo{30.17}{02.77}$). On top of this, we see that the JADE model which learns to jointly disentangle SVHN and MNIST features performs even better than the former methods, sitting at $\oo{29.08}{00.92}$. This is in line with our hypothesis that only the directions of variation shared between MNIST and SVHN (i.e., content) will contribute positively to classification performance on SVHN, and other factors of variation should be disentangled.

We hypothesize that actively attempting to disentangle variation factors (i.e., in JADE) is better than allowing the network to attempt to discard uninformative factors (i.e., paired classifier) given the sample-constrained regime. To assert that the JADE setup is indeed disentangling variation factors, we conduct the following simple experiment: observe the variation in latent space values as different types of samples are passed into the network. In Fig. 2a, we have shown how latent activations change when the SVHN CNN is fed with $500$ samples from the same class (i.e., same content but varying style). These activations are shown for the $20$ latent parameters (of which only $10$ are passed into the MLP classifier, and all used for reconstruction) across $10$ classes of digits in MNIST. We observe that in this setup where content is fixed, the normalized variance of the latent variables that are fed into the MLP classifier is much lower than the variance of latent variables that are solely used for reconstruction. In Fig. 2b, we observe an interesting and complementary phenomena when we pass in $2500$ randomly selected test samples into the SVHN CNN. Here, both the style and the content vary between input samples, and we observe that all $20$ latent parameters are active given the varying input. These observations suggest that JADE is able to successfully disentanglement content and style in low-data SVHN using the help of MNIST as an auxiliary similar dataset.
% the normalized variance activations of 20 latent parameters as 

% \hl{one could argue that because you're doing classification, instead of jointly disentangling,... can't you just jointly discard uninformative directions of variation...}

%% %% %% %% %% %% %% %% %% %% %% %% %% %% %% %% %% %% %% %% %% %% %% %% %% %% %% %% %% %% 
%% %%          CONCLUSION AND FUTURE WORK                                           %% %%
%% %% %% %% %% %% %% %% %% %% %% %% %% %% %% %% %% %% %% %% %% %% %% %% %% %% %% %% %% %% 
\vspace{-.1cm}
\section{Conclusion and Future Work}
\vspace{-.1cm}
In this work, we explore the application of feature disentangling for the problem of supervised classification in a setting where few labeled samples exist, and there are no unlabeled samples for use in unsupervised training. Instead, a similar datasets exists which shares at least one direction of variation with the sample-constrained datasets. We train our model end-to-end using the framework of variational autoencoders and experimentally demonstrated that using a secondary dataset with similar content to SVHN leads to improvements in supervised classification performance.

Given the autoencoding structure of the proposed framework, a reasonable next step is to explore using an ensemble of auxiliary datasets, say one for content and another for style, to augment not only the classification power of the system, but also its reconstruction and generation ability. Currently, reconstruction quality is lacking as samples are being generated using the limited samples. Finally, an exciting extension of the JADE framework is cross-task or cross-modality data synthesis, e.g., learning a joint representation that captures high-level concepts for all modalities of the same object allows for bi-directional generation of missing modalities from the remaining modalities \cite{suzuki2016joint}.

% one can use the joint disentangled embeddings to generate samples with content from MNIST and style from SVHN. This has potential to contribute to the field of Neural Style Transfer.

\vspace{-.2cm}
\small
\bibliography{dis_class}
\bibliographystyle{abbrv}

\end{document}